\documentclass[sigconf]{acmart}

\setcopyright{rightsretained}
\copyrightyear{2026}
\acmYear{2026}
\acmConference[RecSys '26 OARS Workshop]{Online \& Adaptive Recommender System (OARS) Workshop}{September 28, 2026}{Minneapolis, Minnesota, USA}
\acmBooktitle{Proceedings of the OARS Workshop at RecSys 2026, September 28, 2026, Minneapolis, Minnesota, USA}

\usepackage{microtype}
\usepackage{listings}
\usepackage{enumitem}
\usepackage{booktabs}
\usepackage{graphicx}
\usepackage{float}

\begin{document}

\title[LMs for Page-Level Layout Decisions in E-commerce Search]{Language Models for Page-Level Layout Decisions in E-commerce Search}
\subtitle{A Study of Prompt-Based and Representation-Based Methods}

\author{Varun Joshi}
\affiliation{%
  \institution{Walmart Global Tech}
  \city{Hoboken}
  \state{NJ}
  \country{USA}
}
\email{varun.joshi0@walmart.com}

\author{Eva C. Song}
\affiliation{%
  \institution{Walmart Global Tech}
  \city{Hoboken}
  \state{NJ}
  \country{USA}
}
\email{eva.song@walmart.com}

\author{ChengXiang Zhai}
\affiliation{%
  \institution{University of Illinois at Urbana-Champaign}
  \city{Urbana}
  \state{IL}
  \country{USA}
}
\email{czhai@illinois.edu}



\begin{abstract}
E-commerce search pages are critical touchpoints for millions of online shoppers. While traditional search engines return a ranked list of results, modern E-commerce search pages increasingly incorporate recommender system modules -- for example, secondary stacks that surface alternative product groupings at specific positions. When introduced appropriately, secondary stacks can improve user engagement; however, suboptimal placement may disrupt browsing flow and degrade the primary results. Unlike traditional search ranking, where evaluation techniques such as interleaving are well established, evaluating page-level layout changes e.g., when and where to insert a secondary stack remains challenging without costly online A/B testing. To address this, we study offline methods for evaluating whether a given layout decision -- specifically, the inclusion of a secondary stack at a particular position -- is beneficial to users. We investigate language models as scalable evaluators by comparing direct prompt-based, prompt-derived feature, and representation-based methods. Our results show that representation-based approaches consistently outperform prompt-based judging in predicting user engagement, suggesting they provide a reliable foundation for offline layout evaluation in E-commerce search.
\end{abstract}

\maketitle

\section{Introduction}
\begin{figure}[htbp]
\centering
\includegraphics[width=0.75\columnwidth]{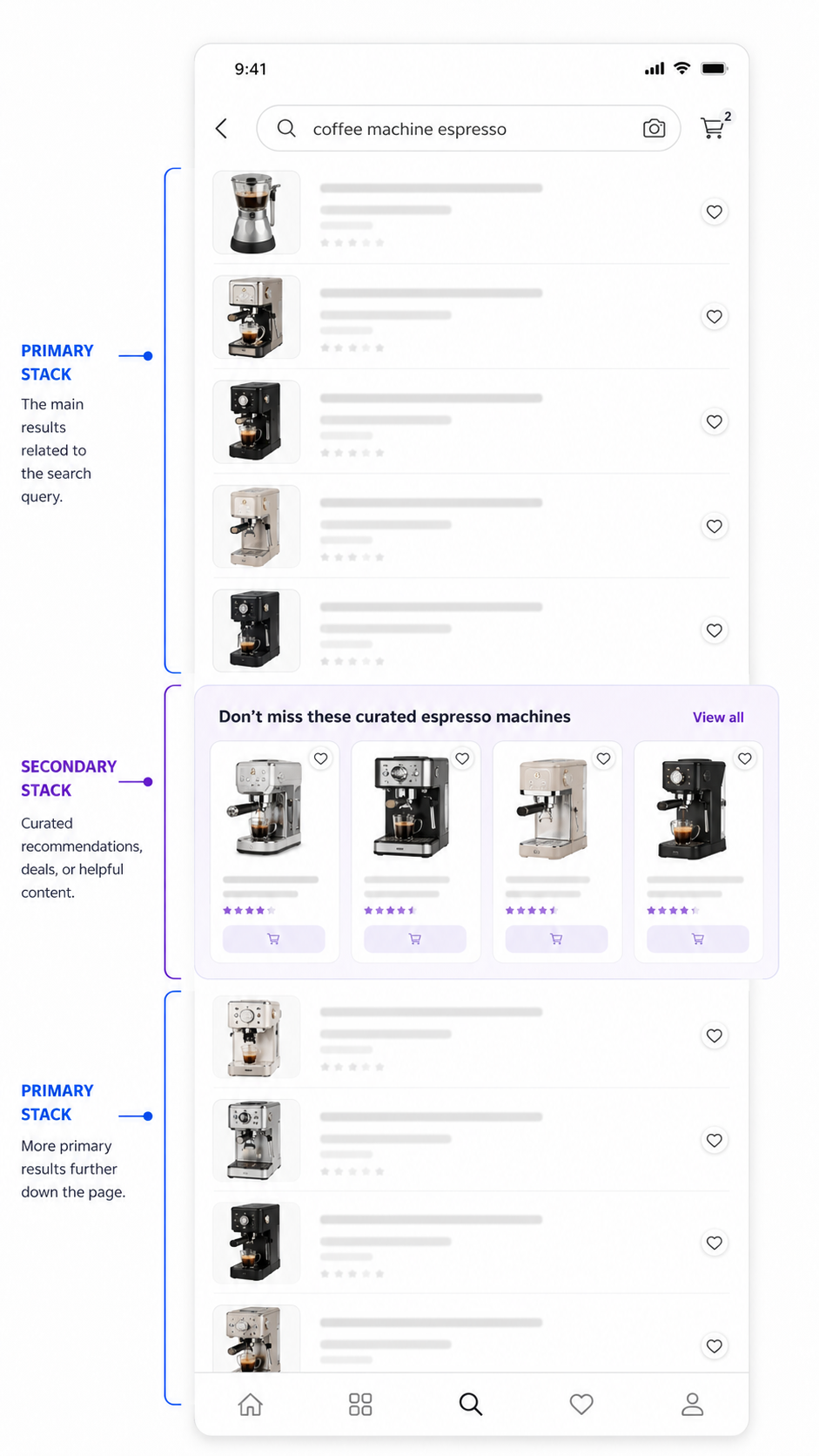}
\Description{Screenshot of an e-commerce search result page on iOS. A horizontal product carousel (secondary stack) is inserted between rows of primary product tiles, surfacing an alternative product grouping at a specific scroll position.}
\caption{Example of a search result page with a secondary stack. The primary stack contains the main ranked results. A secondary stack (horizontal carousel) is inserted at a specific position, surfacing curated product recommendations. Below it, the primary stack continues.}
\label{fig:secondary_stack_example}
\end{figure}

With the growth of online shopping, search engines on E-commerce platforms directly affect the productivity and quality of life of millions of users. While ranking algorithms have been studied extensively, modern search result pages increasingly involve page-level presentation decisions involving recommender modules -- such as whether to insert a secondary stack, and where to place it -- to help users discover alternative intents or facets (see Figure~\ref{fig:secondary_stack_example} for an example). Prior work \cite{song2025practical} has shown that when applied judiciously, such secondary stacks can improve user engagement. However, how to evaluate these layout decisions remains an open challenge. Unlike document ranking, where evaluation methodologies are relatively mature, page layout changes couple multiple components and UI constraints and often require costly online experimentation to assess reliably. Moreover, while historical engagement data provides high-fidelity signals for offline assessment, it is restricted to layouts that have actually been surfaced to users, leaving unseen layouts with no corresponding behavioral signals. These limitations motivate our exploration of language models as a complementary approach for evaluating unseen page layouts offline.

Prior work has studied language models as judges for relevance, but their use for evaluating layout decisions remains unexplored. In this paper, we conduct the first study of using language models to perform offline evaluation of page-level layout decisions in E-commerce search. We formulate this problem as a supervised prediction task grounded in logged user engagement, and conduct a controlled empirical comparison of three classes of approaches: (i) direct prompt-based judging, (ii) prompt-derived feature classifiers, and (iii) representation-based methods. Through systematic experiments and ablations, we analyze the effectiveness of these approaches including the role of explicit page-level features. To our knowledge, this work provides the first unified benchmark comparing these approaches for predicting the behavior impact of page-level layout decisions.

\section{Related Work}
Recent work in information retrieval \cite{rahmani2024llmjudge,rahmani2024llm4eval,mehrdad2024large} studies large language models (LLMs) as judges for relevance assessment, evaluating agreement with human annotations and analyzing reliability and bias \cite{soboroff2025don}. These efforts focus on query-document relevance under structured task settings and do not consider page-level layout decisions with interaction and presentation constraints.

In Human-Computer Interaction (HCI), language models have been used to critique or evaluate user interfaces. Duan et al. generate heuristic-style feedback on UI mockups to support designers \cite{duan2023towards,duan2024generating}, and Luera et al. benchmark multimodal LLMs as UI judges based on alignment with human subjective perception \cite{luera2025mllm}. However, these approaches emphasize design feedback or perception alignment rather than predicting behavioral outcomes from concrete layout decisions using engagement logs.

Language models have also been applied to UI and layout generation or understanding. LayoutPrompter \cite{lin2023layoutprompter} demonstrates constrained layout generation, and related work explores multimodal UI understanding and screen annotation \cite{baechler2024screenai}. While showing that layouts can be represented and manipulated by LLMs, these studies do not evaluate whether specific layout choices improve user engagement, nor compare LLM-based reasoning with simpler representation based predictors trained on historical interaction data.

From an IR perspective, page-level layout decisions relate to interactive retrieval interfaces. Zhang and Zhai's ``card playing" framework models interface presentation as sequential interaction choices \cite{zhang2015information}, but emphasizes formal modeling rather than empirical offline evaluation of layout decisions with large-scale behavioral data.

\section{Problem Formulation}
Given a query $q$, let the primary ranked list of products be $P=(p_1, \ldots, p_n)$, and a candidate secondary stack be $S = (s_1, ..., s_k)$. For a position $i\in[0, n]$, we compare a test layout $(p_1, \ldots, p_i, S, p_{i+1}, \ldots, p_n)$ against the baseline layout $P$ without the secondary stack.

We formulate the problem as binary classification, i.e. with the secondary stack is better (1) v.s. without the secondary stack is better (0). We use Add-to-Cart (ATC) as the engagement signal to indicate positive user experience.

The model is provided with the search context, including the query $q$, secondary stack title $m$, items in $S$, primary results $P$, and key product attributes (e.g. title, brand). This context is supplied to the language models via structured prompts, along with system-level instructions on the UI.

\subsection{Labeling scheme} \label{sec: labeling}
We use historical search log data for evaluation and training. From the search log, we extract records in which a secondary stack was displayed on the search page, and assign labels based on observed Add-to-Cart (ATC) engagement. We adopt a top-to-bottom user scroll model: since primary stack items above the secondary stack position are reached before the user encounters the module, ATCs from those items do not establish that the secondary stack was seen. We therefore consider only ATC events from items at or below the secondary stack position.

Let $\text{ATC}_{S}$ denote whether the user engaged (via ATC) with the secondary stack, and $\text{ATC}_{P^{+}}$ denote whether the user engaged with any primary stack item positioned below the secondary stack. The labeling rule is:
\begin{itemize}
\item $\text{ATC}_{S} = 1 \;\rightarrow$ label $= 1$: the user engaged with the secondary stack, indicating it provided value
\item $\text{ATC}_{S} = 0$,\; $\text{ATC}_{P^{+}} = 1 \;\rightarrow$ label $= 0$: the user scrolled past the secondary stack without engaging, then added to cart from a primary result below --- the secondary stack was seen but did not provide value, and may have imposed an additional scroll cost
\item $\text{ATC}_{S} = 0$,\; $\text{ATC}_{P^{+}} = 0 \;\rightarrow$ discard: we cannot confirm the user scrolled past the secondary stack, so no reliable label can be assigned
\end{itemize}
This scheme ensures that all retained examples correspond to sessions where the user demonstrably encountered the secondary stack, reducing label noise in both training and evaluation.

\section{Experimental Setup}
We study three complementary modeling approaches that differ in how semantic information is represented and used. The first class of approaches leverages Mistral-7B-Instruct-v0.3 language model \cite{jiang2023mistral} to directly judge the usefulness of a given layout through prompting. The second class derives structured features from the same language model output and uses them as inputs to lightweight classifiers. The third class adopts representation-based methods, where a pretrained text encoder -- ModernBERT \cite{warner2025smarter} is used to produce semantic embeddings of the page context using the same prompt, followed by a shallow prediction model. All approaches are trained and evaluated using the same labeling scheme and data splits, enabling a controlled comparison of their effectiveness.

\subsection{Direct Prompting to produce final judgement}
In this setup, we provide the system instruction by explaining the overall task, the use cases of different types of secondary stacks and some potentially helpful general context. A few examples are also provided as sample output. The language model is instructed to output both a binary decision and a score between zero and one.

We considered two variations which use the same prompt (Listing~\ref{lst:prompt}) but differ in the system instruction: a) Direct Prompting (DP); and b) Prompting with Decomposed Judgement (PDJ). PDJ utilizes the enhanced system instruction from Listing~\ref{lst:si-rubric} while DP uses Listing~\ref{lst:si-basic}. PDJ extends DP with additional guidance to decompose the judgement into six intermediate criteria: (1) intent alignment, (2) product relevance, (3) brand popularity, (4) customer value alignment, (5) duplication with primary results, and (6) flow interruption cost. We hypothesize that such a decomposition may have a similar benefit to chain-of-thought prompting. For each criterion, the model scores the layout on a 3-point scale using a provided rubric.

\begin{lstlisting}[caption={Prompt (page context) for the language model}, label={lst:prompt}]
Customer query: $q$
Secondary stack name: $m$
Secondary stack position: $i$
Number of products in the secondary stack: $k$
List of product_ids in the secondary stack: ($s_1, \ldots, s_k$)
List of product titles in the secondary stack: ($t^{(S)}_1, \ldots, t^{(S)}_k$)
List of product brands in the secondary stack: ($b^{(S)}_1, \ldots, b^{(S)}_k$)
List of product values in the secondary stack: ($v^{(S)}_1, \ldots, v^{(S)}_k$)
List of product_ids in the primary stack: ($p_1, \ldots, p_n$)
List of product titles in the primary stack: ($t^{(P)}_1, \ldots, t^{(P)}_n$)
List of product brands in the primary stack: ($b^{(P)}_1, \ldots, b^{(P)}_n$)
List of product values in the primary stack: ($v^{(P)}_1, \ldots, v^{(P)}_n$)
\end{lstlisting}

\begin{lstlisting} [caption={Basic system instruction}, label={lst:si-basic}]
[BACKGROUND] You are an expert at understanding customer behavior on a large-scale e-commerce website. You will be given a pair of layouts for a search results page and your goal is to determine which layout is better. The search results page is for the iOS platform and has a single column for continuous scrolling. Each row contains either a product tile or a horizontal carousel. The list of products in the product tiles are known as the primary stack as they are the main search result. The horizontal carousel also contains a list of products and it is called the secondary stack. The secondary stack horizontal carousel takes the height (real estate space) of roughly 1.5 times of the product tile. The task is to decide whether the search experience is better with the secondary stack or without the secondary stack.

[STACK CONTEXT] Please note that we have the following six secondary stacks:
1. "aaa" stack is ...
...
6. "fff" stack is ...

[POSITION CONTEXT] You will be provided the position of the secondary stack module on the search page. The `Module position: K` means the secondary stack is inserted after primary stack item tile position K. Please consider flow interruption cost and reason about user cognitive state at the scroll depth. Also, consider the quality of primary stack items close in position to the secondary stack. For example, a stack can be helpful at position 12 and annoying at position 4 if the primary stack items at position 5-10 are of very high quality.

[TASK] Layout A is for the search page with the secondary stack module inserted at Secondary stack position. Layout B does not contain the secondary stack. Based on the above information, do you think adding the secondary stack module improves user engagement on the search page? In other words, is layout A better than layout B?

[FORMAT] Your output should be in the following JSON format:
{{
    "final_assessment_score": "<a float number between 0 and 1 indicating the overall quality of Layout A>",
    "conclusion": "<either 'Layout A is better' or 'Layout B is better'>"
}}

[EXAMPLES] Here are a few examples of the expected output:
Example 1:
{{
    "final_assessment_score": "1.0",
    "conclusion": "Layout A is better"
}}
Example 2:
{{
    "final_assessment_score": "0.0",
    "conclusion": "Layout B is better"
}}
Example 3:
{{
    "final_assessment_score": "0.6",
    "conclusion": "Layout A is better"
}}
Example 4:
{{
    "final_assessment_score": "0.2",
    "conclusion": "Layout B is better"
}}
\end{lstlisting}

\begin{lstlisting}[caption={System instruction with rubric}, label={lst:si-rubric}]
[TASK] ...
[RUBRIC] In order to answer this question, please score layout A on each of the following factors:
1. The intent alignment of the secondary stack to the search query (Aligned: 2, Somewhat Aligned: 1, Not Aligned: 0)
2. The relevance of the product titles in the secondary stack to the search query (Relevant: 2, Somewhat Relevant: 1, Not Relevant: 0)
3. The brand popularity of the products in the secondary stack (Highly Popular: 2, Popular: 1, Less Popular: 0)
4. The customer value alignment of the products in the secondary stack (Highly Aligned: 2, Somewhat Aligned: 1, Not Aligned: 0)
5. Whether there is repetition between the products in the secondary stack and the primary stack ...  (Severe Overlap: 2, Acceptable Overlap: 1, No Overlap: 0)
6. Flow interruption cost based on the position of the secondary stack and the comparative quality of nearby primary stack items ... (Low Interruption Cost: 2, Medium Interruption Cost: 1, High Interruption Cost: 0)
[FORMAT] ...
\end{lstlisting}

\subsection{Prompt augmented features + Classification (PFC)}
In this approach, instead of using the final outputs from PDJ directly, we use the decomposed scores as input signals to train simple classification models using the labeled data. Here we use PDJ to generate outputs on an additional training dataset. The labels for the training data are consistent with the held-out test dataset, which is the true engagement based labeling introduced in Section \ref{sec: labeling}.

\subsection{Representation-based embedding + Classification (REC)}
Here, we use the prompt (Listing~\ref{lst:prompt}) as page-level context and input this to the pretrained ModernBERT to generate an embedding for the search page. We then add one hidden layer of $32$ neurons on top of the embedding to form a binary classification model and utilize the same training and held-out test dataset as in PFC.

\subsection{Data Description}
We sample data from one week of traffic on Walmart.com's search result pages which had exactly one secondary stack displayed. Each example includes the query $q$ and various attributes of the primary and secondary stack as described in the previous sections. We assign a binary label to each example using the scheme from Section \ref{sec: labeling}. Our total dataset size consists of 11,000 unique requests which we randomly split into 8,800 (80\%) as the train set and 2,200 (20\%) as the held-out test set to ensure no cross-split leakage. The dataset is balanced to ensure equal distribution among positive and negative labels.

We emphasize that while DP and PDJ require no labeled training data, both PFC and REC involve supervised classifiers trained on the same labeled dataset. The comparison between PFC and REC is therefore fully controlled --- same labels, splits, and downstream classifier architecture --- so the performance gap between them constitutes a controlled comparison between prompt-derived features and pretrained embeddings as input representations. We note that embedding representations could also be extracted directly from Mistral-7B-Instruct-v0.3 model and used for classification. However, since our focus is on comparing prompt-based and representation-based methods, we choose a substantially smaller pretrained encoder to assess whether lightweight embeddings can already match or exceed prompt-based methods.

\section{Results and Discussion}
\begin{figure}[!htbp]
\centering
\includegraphics[width=\columnwidth]{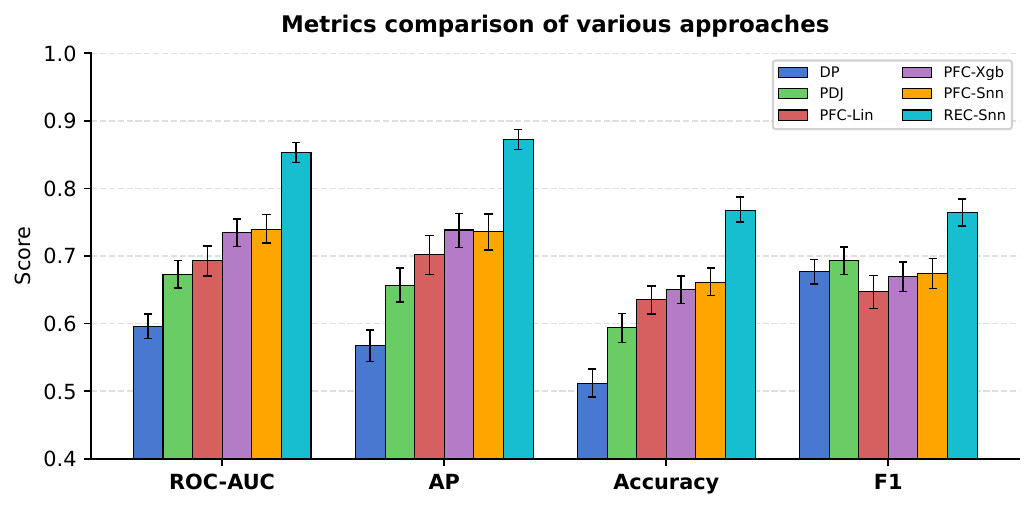}
\Description{Bar chart comparing ROC-AUC, Average Precision, Accuracy, and F1 metrics for six approaches: DP, PDJ, PFC-Lin, PFC-Xgb, PFC-Snn, and REC-Snn, with 95\% bootstrap confidence intervals. REC-Snn achieves the highest scores across all metrics.}
\caption{Metrics comparison of various approaches with 95\% bootstrap confidence intervals. ``-Lin", ``-Xgb" and ``-Snn" denote binary classification using a linear model, XGB model, and shallow neural network, respectively (refer Sections 4.2 and 4.3). Full numerical results are provided in Table~\ref{tab: approach_comparison}.}
\label{fig: approach_comparison}
\end{figure}

We present the mean and 95\% bootstrap confidence intervals for various binary classification metrics for the three classes of approaches in Figure~\ref{fig: approach_comparison} on the held-out test set. For each variation, we report ROC-AUC, Average Precision (AP), Accuracy at threshold $0.5$, and F1 score at threshold $0.5$. It can be seen that the variation of using representation-based embedding plus shallow network as classifier dominates across all the metrics.

\begin{table*}[t]
\centering
\small
\caption{Metrics comparison of various approaches with 95\% bootstrap confidence intervals}
\label{tab: approach_comparison}
\begin{tabular}{lcccc}
\toprule
Approach & ROC-AUC & AP & Accuracy & F1 \\
\midrule
DP & 0.5961 {\scriptsize(0.5778--0.6144)} & 0.5673 {\scriptsize(0.5442--0.5904)} & 0.5120 {\scriptsize(0.4914--0.5327)} & 0.6772 {\scriptsize(0.6589--0.6951)} \\
PDJ & 0.6730 {\scriptsize(0.6530--0.6937)} & 0.6564 {\scriptsize(0.6317--0.6821)} & 0.5943 {\scriptsize(0.5718--0.6150)} & 0.6933 {\scriptsize(0.6723--0.7132)} \\
PFC-Lin & 0.6927 {\scriptsize(0.6702--0.7147)} & 0.7022 {\scriptsize(0.6728--0.7303)} & 0.6351 {\scriptsize(0.6145--0.6559)} & 0.6479 {\scriptsize(0.6224--0.6710)} \\
PFC-Xgb & 0.7343 {\scriptsize(0.7142--0.7548)} & 0.7385 {\scriptsize(0.7129--0.7631)} & 0.6504 {\scriptsize(0.6295--0.6705)} & 0.6703 {\scriptsize(0.6473--0.6913)} \\
PFC-Snn & 0.7396 {\scriptsize(0.7192--0.7615)} & 0.7361 {\scriptsize(0.7092--0.7623)} & 0.6615 {\scriptsize(0.6414--0.6823)} & 0.6742 {\scriptsize(0.6519--0.6965)} \\
REC-Snn & \textbf{0.8530} {\scriptsize(0.8381--0.8683)} & \textbf{0.8728} {\scriptsize(0.8574--0.8872)} & \textbf{0.7680} {\scriptsize(0.7505--0.7873)} & \textbf{0.7645} {\scriptsize(0.7444--0.7841)} \\
\bottomrule
\end{tabular}

\vspace{1mm}
\footnotesize
\textit{Note.} ``-Lin", ``-Xgb" and ``-Snn" denote the corresponding method combined with binary classification using a linear model, XGB model, and shallow neural network model, respectively. Values in parentheses are 95\% bootstrap confidence intervals.
\end{table*}

We first analyze the errors produced by DP and PDJ. Recall that the models output both a binary verdict and a float score between 0 and 1. Results in Figure~\ref{fig: approach_comparison} are based on the float score. We note that DP labeled every single instance as 1, i.e. always favored showing the secondary stack. Further, despite few-shot examples spanning the full [0,1] range, DP only outputs 0.7 or 0.8. On the contrary, PDJ produces 12\% negative binary decisions and outputs a wider range of float scores. This suggests that the decomposition based chain-of-thought helps the overall decision.

To better understand how PDJ utilized the dimensional decomposed scores, we fit a linear regression model by using the decomposed scores as features and the real-valued output score as the label. The result shows equal weights among all the features. This indicates that the final verdict provided by the language model is simply a combination of the score from each dimension equally. While PDJ improves over DP, the equal weighting suggests the language model does not optimally combine the decomposed scores, motivating the use of explicit classifiers.

This hypothesis is confirmed by the improvement in metrics we see for the class of PFC-X models (e.g., see PFC-Xgb vs PDJ in Figure~\ref{fig: approach_comparison}). Note that this group of results is obtained with all the decomposition scores generated by the prompt PDJ and a few "emphasized" additional features, e.g. the secondary stack type and the position on the page. These features are already present in the prompt but are added back explicitly to emphasize their importance. Next, we examine the feature importance for this class of models. The top three features across all of the variations are ``flow interruption cost" (6), ``repetition level" (5), and ``intent alignment" (1). Upon investigating the coefficients of the linear models, we find that all the features have positive weights except ``repetition level". This is intuitive: high duplication causes browsing fatigue and provides little information gain over existing primary results. The negative coefficient for ``repetition level" was not reflected in PDJ's equal weighting, further indicating that the language model was not able to appropriately combine the decomposed features.

Importantly, PFC and REC are trained under identical conditions --- same labels, splits, and downstream classifier architecture --- so the performance gap between them constitutes a controlled comparison between prompt-derived features and pretrained embeddings as input representations. As shown in Figure~\ref{fig: approach_comparison}, the representation based approach significantly outperforms (as measured by 95\% bootstrap confidence intervals) all prompt-based methods across all the metrics, including PFC which also involves supervised training. A possible explanation is that the task primarily requires recognizing semantic patterns in page context that correlate with historical engagement, rather than explicit reasoning about interface design principles. The frozen text encoder implicitly aggregates multiple signals, such as topical alignment and redundancy, into a continuous representation that can be effectively exploited by a simple classifier. In contrast, prompt-based reasoning introduces additional variance and relies on discrete intermediate outputs that may not align with the label.

\subsection{Ablation study}
\begin{figure}[!htbp]
\centering
\includegraphics[width=\columnwidth]{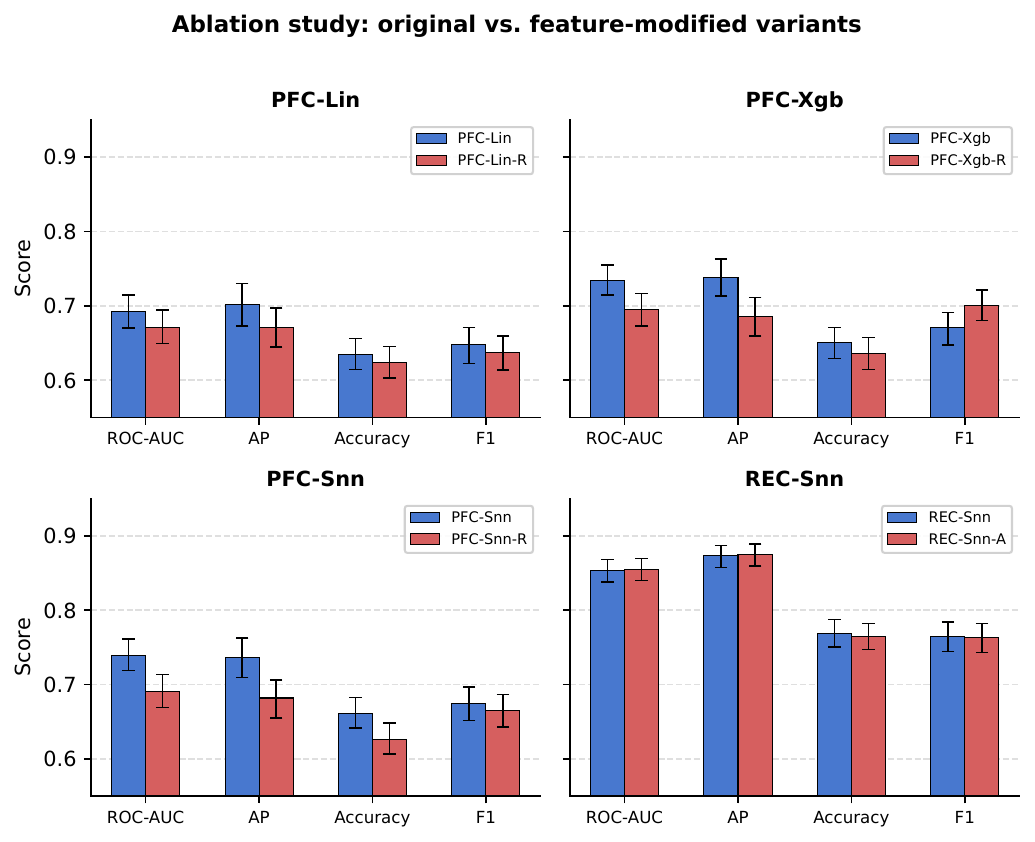}
\Description{Bar chart showing ablation study results. PFC models (Lin, Xgb, Snn) show AUC drops of 0.02 to 0.05 when explicit features are removed. REC-Snn shows negligible AUC change when explicit features are added, indicating the embedding already encodes those signals.}
\caption{Ablation study comparing each model (blue) against its feature-modified variant (red) with 95\% bootstrap confidence intervals. For PFC models, ``-R" removes explicit features; for REC-Snn, ``-A" adds them. Explicit features include secondary stack type, insertion position, stack size, and duplicated products count. Full numerical results are provided in Table~\ref{tab:ablation}.}
\label{fig:ablation}
\end{figure}

To better understand the sources of predictive signal across different approaches, we conduct an ablation study focusing on the role of explicit page-level features. The results of the study are shown in Figure~\ref{fig:ablation}. When using only PDJ-derived judging scores as inputs to a classifier, we observe substantially lower performance. However, augmenting these scores with a small set of explicit features that are already present in the prompt, such as secondary stack type and insertion position, leads to a notable improvement. In contrast, for the representation-based approach, adding the same explicit features on top of pretrained text embeddings yields no measurable improvement. As shown in Figure~\ref{fig:ablation}, removing explicit features consistently degrades PFC models (AUC drops of 0.02--0.05), while adding them to REC-Snn yields negligible change (AUC $+0.002$), indicating that the embedding already encodes these signals.

\begin{table*}[t]
\centering
\small
\caption{Ablation study of features with 95\% bootstrap confidence intervals}
\label{tab:ablation}
\begin{tabular}{lcccc}
\toprule
Approach & ROC-AUC & AP & Accuracy & F1 \\
\midrule
PFC-Lin-R & 0.6713 {\scriptsize(0.6496--0.6946)} & 0.6705 {\scriptsize(0.6444--0.6969)} & 0.6242 {\scriptsize(0.6027--0.6450)} & 0.6371 {\scriptsize(0.6137--0.6591)} \\
PFC-Xgb-R & 0.6948 {\scriptsize(0.6727--0.7166)} & 0.6850 {\scriptsize(0.6595--0.7108)} & 0.6353 {\scriptsize(0.6145--0.6573)} & 0.7007 {\scriptsize(0.6802--0.7214)} \\
PFC-Snn-R & 0.6909 {\scriptsize(0.6692--0.7133)} & 0.6819 {\scriptsize(0.6548--0.7063)} & 0.6265 {\scriptsize(0.6068--0.6482)} & 0.6647 {\scriptsize(0.6428--0.6863)} \\
REC-Snn-A & 0.8550 {\scriptsize(0.8398--0.8697)} & 0.8742 {\scriptsize(0.8590--0.8888)} & 0.7649 {\scriptsize(0.7468--0.7823)} & 0.7637 {\scriptsize(0.7428--0.7821)} \\
\bottomrule
\end{tabular}

\vspace{1mm}
\footnotesize
\textit{Note.} ``-R" indicates the original model presented in Table~\ref{tab: approach_comparison} with explicit features removed and ``-A" indicates the original model with explicit features added. Explicit features include secondary stack type, insertion position, stack size, and duplicated products count. Values in parentheses are 95\% bootstrap confidence intervals.
\end{table*}

\section{Conclusion}
In this work, we conducted the first study of offline evaluation methods for assessing page-level layout decisions on E-commerce search result pages, with a focus on the inclusion of secondary stacks at specific positions. We compared prompt-based judging, prompt-derived feature classifiers, and representation-based methods. Across all experiments, the representation-based method achieved stronger predictive performance. An ablation analysis further showed that prompt-based processing may be lossy for structured attributes, whereas representation-based models capture such information more efficiently. Taken together, these findings indicate that, within the scope of our study, lightweight semantic representations provide a reliable foundation for offline page-level layout evaluation.

\bibliographystyle{ACM-Reference-Format}
\bibliography{references}

\end{document}